\documentclass[10pt, a4paper]{article}

\usepackage{lrec-coling2024} 

\usepackage[utf8]{inputenc}
\usepackage{graphicx}
\usepackage{diagbox}
\usepackage{xcolor,colortbl}
\usepackage{soulutf8}
\usepackage[autolanguage]{numprint}
\usepackage{xspace}
\usepackage{booktabs}
\usepackage{paralist}

\title{A Benchmark Evaluation of \\Clinical Named Entity Recognition in French}

\name{Nesrine Bannour$^1$, Christophe Servan$^1$, Aurélie Névéol$^1$, Xavier Tannier$^2$} 

\address{$^1$Université Paris-Saclay -- CNRS -- LISN, \\$^2$Sorbonne Université --  Inserm -- Université Sorbonne Paris Nord -- LIMICS \\
         Paris, France \\
         $^1$\{firstname.lastname\}@lisn.upsaclay.fr, $^2$\{firstname.lastname\}@sorbonne-universite.fr\\}

\abstract{
\textbf{Background:}
Transformer-based language models have shown strong performance on many Natural Language Processing (NLP) tasks. Masked Language Models (MLMs) attract sustained interest because they can be adapted to different languages and sub-domains through training or fine-tuning on specific corpora while remaining lighter than modern Large Language Models (LLMs). Recently, several MLMs have been released for the biomedical domain in French, and experiments suggest that they outperform standard French counterparts. However, no systematic evaluation comparing all models on the same corpora is available.   
\textbf{Objective:}
This paper presents an evaluation of masked language models for biomedical French on the task of clinical named entity recognition. 
\textbf{Material and methods:}
We evaluate biomedical models \texttt{CamemBERT-bio} and \texttt{DrBERT} and compare them to standard French models \texttt{CamemBERT}, \texttt{FlauBERT} and \texttt{FrALBERT} as well as multilingual \texttt{mBERT} using three publically available corpora for clinical named entity recognition in French. The evaluation set-up relies on gold-standard corpora as released by the corpus developers.  
\textbf{Results:}
Results suggest that \texttt{CamemBERT-bio} outperforms \texttt{DrBERT} consistently while \texttt{FlauBERT} offers competitive performance and \texttt{FrAlBERT} achieves the lowest carbon footprint. 
\textbf{Conclusion:}
This is the first benchmark evaluation of biomedical masked language models for French clinical entity recognition that compares model performance consistently on nested entity recognition using metrics covering performance and environmental impact. 
 \\ \newline  }

\begin{document}

\maketitleabstract

\section{Introduction}

The recent development of Clinical Data Warehouses in many French hospitals ~\cite{Jannot2017georges,madec2019ehop,pressat2022evaluation} is making unstructured clinical data, including narratives, available for secondary use. As a result, there is a growing need in the biomedical community for Natural Language Processing tools that facilitate the extraction of clinical information from text to support epidemiological studies. Many epidemiological indicators can be modeled as named entities to be extracted from the raw text of clinical reports.  

For this reason, the task of Named Entity Recognition, or NER, has attracted a lot of attention in the past decades, in particular through shared tasks aiming at direct comparison of methods. Earlier challenges offered tasks for English~\cite{uzuner2007evaluating,uzuner20112010} but more recently, other languages have also been addressed~\cite{marimon2019automatic,intxaurrondo2018finding}, including French~\cite{neveol2015clef,cardon-etal-2020-presentation}. 

The evaluation resources released in the shared tasks continue to be used for evaluating new methods and tools. However, individual efforts often come with adaptations of the data sets or metrics so that comparability is not possible across the board. 

Herein, we address this issue by presenting a systematic evaluation that offers comparability across systems as well as with the literature introducing the reference corpora. 
 The main contributions of this paper are: 
\begin{itemize}
    \item A benchmark evaluation of clinical named entity recognition in French based on original gold standard annotations, including nested entities
    \item A comparison of freely available masked language models for general and biomedical French on the NER task
    \item A comparison to strong symbolic baselines
 \end{itemize}
 
\section{Corpora}
This section briefly presents the clinical French corpora used to train NER models and evaluate the systems considered in this benchmark. 
\begin{itemize}
    \item \textbf{DEFT}~\cite{cardon-etal-2020-presentation} is a subset of 167 clinical cases from the CAS corpus~\citeplanguageresource{grabar-etal-2018-cas}, introduced in the DEFT challenge in 2020\footnote{\url{https://deft.limsi.fr/2020/index-en.html.}}. This corpus is annotated with 13 types of clinical entities and five attributes. It is divided into a training set of 85 documents, a validation set of 20 documents, and a test set of 62 documents.
    
    \item \textbf{E3C}~\citeplanguageresource{magnini2021e3c} is a European corpus of clinical cases. We use the French subcorpus, which comprises 1,615 clinical cases collected in the public domain. 
    It is annotated with 6 types of named clinical entities,
    including \texttt{CLINENTITY}, which we disaggregate into subtypes in order to have an annotation scheme with a diversity approaching that of other corpora. Each entity of this type is associated with a Concept Unique Identifier from the UMLS (Unified Medical Language System) metathesaurus, which can be used to retrieve semantic groups \cite{mccray2001aggregating}. 
    In our experiments, we use the gold-standard annotations in the first layer.  We use 20\% of the training set for validation in the NER models. 
    
    \item \textbf{QUAERO French Med}~\citeplanguageresource{neveol2014quaero} comprises documents belonging to two text genres, which we treat separately. The EMEA subcorpus is a collection of 13 patient information leaflets supplied by the European Medicines Agency that describes drugs marketed in Europe. The MEDLINE subcorpus consists of 2,500 titles of scientific articles indexed in the MEDLINE database\footnote{\url{http://pubmed.ncbi.nlm.nih.gov/}}. The entire corpus is annotated with 10 types of clinical entities derived from UMLS semantic groups.
    
\end{itemize}

\begin{figure}[t]
\centering
\includegraphics[width=0.35\textwidth]{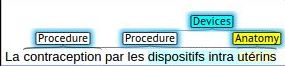}
\caption{3 layers of nested entities in an excerpt of the MEDLINE corpus \label{QUAERO_FrenchMed_MEDLINE}}
\end{figure}

Table~\ref{stats_medical} presents general descriptive statistics of the study corpora. 
\begin{table}[h]
\centering
\begin{tabular}{lrrr}
\toprule
 & \textbf{train} & \textbf{dev} & \textbf{test} \\
\midrule
&\multicolumn{3}{c}{\textit{DEFT}} \\
\textbf{Tokens} & \numprint{31752} & \numprint{5076} & \numprint{20360} \\
\textbf{Entities (all)} & \numprint{7584} & \numprint{1432} & \numprint{5140} \\
 \textbf{Entities (Unique)} & \numprint{5037} & \numprint{1230}  & \numprint{3809} \\
\midrule
&\multicolumn{3}{c}{\textit{E3C$_{FR}$}} \\
\textbf{Tokens} & \numprint{19808} & - & \numprint{4671} \\
\textbf{Entities (all)} & \numprint{3406} & - & \numprint{706} \\
\textbf{Entities (Unique)} & \numprint{2197} & - & \numprint{566} \\
\midrule
&\multicolumn{3}{c}{\textit{EMEA}} \\
\textbf{Tokens} & \numprint{14944} & \numprint{13271} & \numprint{12042} \\
\textbf{Entities (all)} & \numprint{2695} & \numprint{2260} & \numprint{2204} \\
\textbf{Entities (Unique)} & \numprint{923} &  \numprint{756} & \numprint{658} \\
\midrule
&\multicolumn{3}{c}{\textit{MEDLINE}} \\ 
\textbf{Tokens} & \numprint{10552} & \numprint{10503} & \numprint{10871} \\  
\textbf{Entities (all)} & \numprint{2994} & \numprint{2977} & \numprint{3103} \\
\textbf{Entities (Unique)} & \numprint{2296} &  \numprint{2288} & \numprint{2390} \\
\bottomrule
\end{tabular}
\caption{Number of tokens and entity annotations in each split of the study corpora.}
\label{stats_medical}
\end{table}
Table~\ref{stats_layers} presents descriptive statistics of the distribution of entities in layers in the study corpora. 
\begin{table}[h]
\centering
\begin{tabular}{rrrr}
\toprule
 \textbf{Layer 1} & \textbf{Layer 2} & \textbf{Layer 3} & \textbf{Layer 4} \\
\midrule
\multicolumn{4}{c}{\textit{DEFT}} \\
65.13\%	&30.84\%	&4.02\%	&0\% \\
\midrule
\multicolumn{4}{c}{\textit{E3C$_{FR}$}} \\
94.47\%	&5.36\%	&0.12\%	&0.06\% \\

\midrule
\multicolumn{4}{c}{\textit{EMEA}} \\
85.73\%	&13.60\%	&0.64\%	&0.03\% \\
\midrule
\multicolumn{4}{c}{\textit{MEDLINE}} \\ 
74.69\%	&23.75\%	&1.50\%	&0.06\% \\
\bottomrule
\end{tabular}
\caption{Distribution of entity annotations in each layer of the study corpora.}
\label{stats_layers}
\end{table}
Figure \ref{QUAERO_FrenchMed_MEDLINE} present an excerpt of the MEDLINE corpus containing annotations over three layers: on Layer 1, "contraception par les dispositifs intra utérins" (\textit{contraception with intra uterine devices}) is annotated with the entity type "PROCEDURE", on layer 2 "dispositifs intra utérins" (\textit{intra uterine devices}) is annotated with the entity type "DEVICE" while "contraception" is annotated as a "PROCEDURE" and on layer 3, "utérins" (\textit{uterine}) is annotated with the entity type "ANATOMY". 

\section{Named Entity Recognition models}
We trained named entity recognition (NER) models using the python library \texttt{NLstruct}~\cite{wajsburtPhD2021}\footnote{\url{https://github.com/percevalw/nlstruct}}. \texttt{NLstruct} NER models comprise a text encoder, a word tagger, and a bounds matcher.
In addition to state-of-the-art performance on NER tasks, \texttt{NLstruct} features the ability to address nested entities, which can be found in QUAERO French Med.  \texttt{NLstruct} also supports annotations in the BRAT standoff format\footnote{The Brat Rapid Annotation Tool - BRAT ~\cite{StenetorpEACLdemo2012} produces annotations in the so-called standoff format described in the BRAT online manual \url{https://brat.nlplab.org/standoff.html}}, which is used by all three corpora in our study.  

The text encoder component in \texttt{NLstruct} relies on embeddings produced by a BERT language model, a char-CNN encoder, and static French FastText embeddings. To compare French MLMs, we train one NER model using each of the language models in our benchmark. These MLMs are available in the HuggingFace transformers library~\cite{wolf-etal-2020-transformers}. However, it can be noted that some level of adaptation was needed at the tokenization step to use some of the models within \texttt{NLstruct}, especially for \texttt{frALBERT} and \texttt{FlauBERT} models\footnote{The source code is available here: \url{https://gitlab.lisn.upsaclay.fr/nlp/deep-learning/nlstruct}}.

\paragraph{General models.} We used the major French MLMs freely available for our experiments and a multilingual model. 
\begin{itemize}
    \item CamemBERT~\cite{martin-etal-2020-camembert}: a language model for French based on the RoBERTa~\cite{liu2019roberta} model that was pretrained on the OSCAR French corpus~\citeplanguageresource{suarez2019asynchronous}. We use the \texttt{camembert-base} model.
    \item FlauBERT~\cite{le2020flaubert}: a language model for French based on the BERT~\cite{devlin-etal-2019-bert} model that was pretrained on a large multiple-source French corpus. We use the \texttt{flaubert-base-uncased} model. 
    \item FrALBERT~\cite{cattan-etal-2021-usability}: a compact model based on the ALBERT~\cite{Lan2020} model that was pretrained on 4GB of French Wikipedia part. We use the \texttt{fralbert-base} model. 
    \item Multilingual BERT~\cite{devlin-etal-2019-bert}: a language model that was pretrained on the 102 languages with the largest Wikipedia, including French. We use the \texttt{bert-base-multilingual-uncased} model. We denote Multilingual BERT as mBERT.
     
\end{itemize}

\paragraph{Domain-specific models.} We used the  French MLMs dedicated to the biomedical domain that were freely available for our experiments. 
\begin{itemize}
    \item CamemBERT-bio~\cite{touchent2023camembert}: an adapted CamemBERT model for the biomedical domain that was built using continual-pretraining from \texttt{camembert-base} model and trained on a created French biomedical corpus from three sources: The ISTEX database, the CLEAR corpus~\citeplanguageresource{grabar-cardon-2018-clear} and the third unannotated layer of the E3C corpus~\citeplanguageresource{magnini2021e3c}. we use the \texttt{camembert-bio-base} model.
    \item DrBERT~\cite{labrak-etal-2023-drbert}: a RoBERTa-based French biomedical model that was trained from scratch on a 
    web-based medical corpus. 
    We use the \texttt{DrBERT-4GB} model.
\end{itemize}
We also considered \texttt{AliBERT}~\cite{berhe-etal-2023-alibert} but did not find a publicly accessible version. Similarly, the model described by~\citep{le-clercq-de-lannoy-etal-2022-strategies} was not available to us. 
\paragraph{Baseline.} As a baseline, we use a symbolic method that builds a dictionary of entities found in the training and development splits of a corpus and simply matches these entities in the test split. In practice, we use the BRAT propagation tool introduced by \citep{grouin-2016-controlled} to "propagate" BRAT annotations to the test sets\footnote{\url{https://github.com/grouin/propa}}.  

\section{Evaluation metrics}
We evaluate the performance of our models at the entity level by measuring the micro Precision, Recall, and F-measure. Confidence intervals at 95\% confidence level were computed using the empirical bootstrap method~\cite[p. 275]{Dekking2007}. Each test corpus is sampled with replacement 1000 times, and evaluation metrics are calculated for each sample.
Baseline scores were computed using \texttt{brateval}~\cite{verspoor2013annotating}. To measure the carbon footprint of training and testing our models, we use the \texttt{Carbon tracker} tool~\cite{anthony2020carbontracker}. These estimates are approximative and are computed by using an average carbon intensity of 58.48 gCO$_2$/kWh corresponding to our location in France. 

\section{Results and discussion}

Tables \ref{results-DEFT} to \ref{results-EMEA} present the results of our NER experiments on the study corpora.

\begin{table*}[!ht]
\centering
\begin{tabular}{lllll}
\hline
\textbf{Models} & \multicolumn{4}{c}{\textbf{DEFT}}\\
& \textit{Precision} & \textit{Recall} & \textit{F-measure} & \textit{CO$_2$ eq (g.)}\\
\hline
CamemBERT & 0.73 [0.71-0.75] & 0.75 [0.73-0.77] & 0.74 [0.73-0.76] & 7.7\\
FlauBERT & 0.73 [0.72-0.76] & 0.75 [0.73-0.77] & 0.74 [0.73-0.76] & 8\\
mBERT & 0.72 [0.70-0.74] & 0.74 [0.72-0.76] & 0.73 [0.71-0.75] & 8.3\\
frALBERT & 0.68 [0.66-0.69] & 0.67 [0.65-0.69] & 0.67 [0.66-0.69]  & 4.5\\
\hline
CamemBERT-bio & \textbf{0.75} [0.73-0.77] & \textbf{0.77} [0.75-0.78] & \textbf{0.76} [0.74-0.78] & 8.7\\
DrBERT & 0.71 [0.68-0.73] & 0.72 [0.70-0.74] & 0.71 [0.69-0.73] & 5.5\\
\hline
Baseline & 0.38	& 0.32&	0.35 & -\\
\citet{copara-etal-2020-contextualized}& -& -& 0.73&-\\
\hline
\end{tabular}
\caption{\label{results-DEFT} Performance of nested entity extraction on the DEFT test set.}
\end{table*}

\begin{table*}[!ht]
\centering
\begin{tabular}{lllll}
\hline
\textbf{Models} & \multicolumn{4}{c}{\textbf{E3C}}\\
& \textit{Precision} & \textit{Recall} & \textit{F-measure} & \textit{CO$_2$ eq (g.)}\\
\hline
CamemBERT & 0.52 [0.42-0.63] & 0.50 [0.45-0.56] & 0.51 [0.46-0.56] & 3.6\\
FlauBERT & \textbf{0.54} [0.49-0.60] & \textbf{0.53} [0.46-0.60] & \textbf{0.54} [0.51-0.57] & 4.1\\
mBERT & 0.51 [0.45-0.58] & 0.52 [0.47-0.57] & 0.52 [0.48-0.54] & 4.8\\
frALBERT & 0.50 [0.41-0.59] & 0.55 [0.51-0.59] & 0.52 [0.47-0.58]  & 2.5\\
\hline
CamemBERT-bio & 0.52 [0.44-0.61] & 0.52 [0.45-0.59] & 0.52 [0.48-0.55] & 3.6\\
DrBERT & 0.47 [0.40-0.57] & 0.52 [0.46-0.60] & 0.49 [0.46-0.53] & 3.6\\
\hline
Baseline & 0.24 &	0.37 &	0.29 & -\\
\hline
\end{tabular}
\caption{\label{results-E3C} Performance of nested entity extraction on the E3C test set.}
\end{table*}

\begin{table*}[!ht]
\centering
\begin{tabular}{lllll}
\hline
\textbf{Models} & \multicolumn{4}{c}{\textbf{MEDLINE}}\\
& \textit{Precision} & \textit{Recall} & \textit{F-measure} & \textit{CO$_2$ eq (g.)}\\
\hline
CamemBERT & 0.64 [0.62-0.66] & 0.66 [0.64-0.67] & 0.65 [0.63-0.66] & 1.9\\
FlauBERT & 0.67 [0.65-0.68] & 0.69 [0.67-0.71] & 0.68 [0.66-0.69] & 2.2\\
mBERT & 0.63 [0.61-0.65] & 0.67 [0.65-0.69] & 0.65 [0.63-0.67] & 3\\
frALBERT & 0.53 [0.51-0.54] & 0.52 [0.49-0.54] & 0.52 [0.50-0.54] & 1.1\\
\hline
CamemBERT-bio & 0.66 [0.65-0.68] & 0.70 [0.68-0.72] & 0.68 [0.66-0.70] & 2.2\\
DrBERT & 0.63 [0.61-0.65] & 0.65 [0.63-0.67] & 0.64 [0.62-0.66] & 2\\
\hline
Baseline & \textbf{0.73}	& 0.30 &	0.42  & -\\
\citet{erasmusCLEFeHealth2016}& 0.68 & \textbf{0.72} & \textbf{0.70} &-\\
\hline
\end{tabular}
\caption{\label{results-MEDLINE} Performance of nested entity extraction on the MEDLINE test set.}
\end{table*}

\begin{table*}[!ht]
\centering
\begin{tabular}{lllllll}
\hline
\textbf{Models} & \multicolumn{4}{c}{\textbf{EMEA}}\\
& \textit{Precision} & \textit{Recall} & \textit{F-measure} & \textit{CO$_2$ eq (g.)}\\
\hline
CamemBERT & 0.66 [0.62-0.70] & 0.65 [0.56-0.73] & 0.65 [0.59-0.72] & 3.9\\
FlauBERT & 0.69 [0.67-0.72] & 0.66 [0.59-0.73] & 0.68 [0.63-0.71] & 4.4\\
mBERT & 0.67 [0.64-0.72] & 0.67 [0.61-0.73] & 0.67 [0.63-0.72] & 4\\
frALBERT & 0.62 [0.57-0.67] & 0.65 [0.61-0.70] & 0.63 [0.59-0.68]  & 2.4\\
\hline
CamemBERT-bio & 0.70 [0.66-0.74] & 0.68 [0.61-0.75] & 0.69 [0.63-0.74] & 5.2\\
DrBERT & 0.69 [0.66-0.72] & 0.64 [0.58-0.71] & 0.66 [0.62-0.71] & 3\\
\hline
Baseline & \textbf{0.73} &	0.43 &	0.55 & -\\
\citet{erasmusCLEFeHealth2016}& 0.72 & \textbf{0.79} & \textbf{0.75} & -\\
\hline
\end{tabular}
\caption{\label{results-EMEA} Performance of nested entity extraction on the EMEA test set.}
\end{table*}

\subsection{NER performance}


\paragraph{Overall entity extraction performance.} The NER models trained using masked language models outperform the symbolic baseline by at least 10 points of F-measure (up to 40 points for DEFT), although the symbolic baseline can exhibit high precision (e.g., on EMEA, MEDLINE). Interestingly, the knowledge-based approach proposed by \citet{erasmusCLEFeHealth2016} continues to achieve the best results on the MEDLINE and EMEA corpora, with an F-measure of 0.7 and 0.75 respectively.

The performance of the general French and multilingual models is quite similar, including the multilingual model\footnote{\citet{copara-etal-2020-contextualized} found \texttt{mBERT} to be outperformed by \texttt{CamemBERT-large}}, and, on average, lower compared to the performance of the biomedical models, except for the E3C corpus, where the \texttt{FlauBERT model} performs better. The \texttt{CamemBERT-bio} biomedical model seems to perform better than the \texttt{DrBERT} biomedical model, suggesting that continual-pretraining from an existing French model on biomedical data might be beneficial in achieving good outcomes. 


\paragraph{Nested entity extraction performance.} A layer-by-layer evaluation would be difficult to perform, due to the difficulty of aligning system outputs and reference annotations at the layer level. However, we can report that the system outputs contain annotations with depth 3 or 4 depending on the specific models and corpora and exhibit a distribution of annotations across layers that is similar to that of reference annotations. This suggests that the nesting of annotations by NLStruct is successful. Moreover, F-measure for the DEFT corpus exceed .65, which would be the ceiling score for a system performing flat-entity extraction of layer 1 entities. This also suggests that the nested entity extraction is performed successfully.   

\paragraph{Entity extraction performance per entity type.} Due to space constraints, we are not providing detailed performance per entity type over the study corpora. Nonetheless, we can notice that the performance of the models tends to vary following similar trends, with highest performance reached for entity types with either high support in terms of training instances and/or high regularity in their occurrence patterns (e.g., temporal entities).

\subsection{Comparability of models and experiments}

To evaluate their biomedical \texttt{CamemBERT-bio} model, \citet{touchent2023camembert} fine-tuned a NER model on the semi-annotated layer 2 of the E3C corpus and evaluated it on the first layer of this corpus, yielding an F-measure of 69.85. A direct comparison with our results is not possible since we train and evaluate our model using only the first layer containing the gold standard annotations. However, it suggests that silver standard annotations can be useful for training an NER model. 

\citet{touchent2023camembert} and \citet{labrak-etal-2023-drbert} evaluated their biomedical MLMs on the two subcorpora of QUAERO French Med and compared them to general French models. However, the task was cast as direct token classification and did not address the nested named entities. Indeed, \citet{touchent2023camembert} removed nested entities by keeping only the coarse entities, whereas \citet{labrak-etal-2023-drbert} concatenated the names of the nested entities to produce new entities and evaluated their results at the token level. These experiments can be seen as intrinsic evaluations of the masked language models. 

In contrast, our experiments aim to address the entity recognition task in an extrinsic setting. Our results suggest that the size training data available to train the NER models
had more impact on NER performance than the language models used: performance of all approaches is generally lower for E3C vs. other corpora. Similarly, when looking at performance on individual entity types, we generally note that entities with the highest prevalence in the training sets yield higher performance.    

\subsection{Carbon footprint}
Tables \ref{results-DEFT} to \ref{results-EMEA} show the carbon footprint of our NER experiments in terms of CO$_2$ equivalent measure in grams.
The highest CO$_2$ carbon emissions are observed when training and testing the DEFT NER models. This is partly due to the fact that this corpus has more tokens than the other corpora, as illustrated in Table \ref{stats_medical}. 
Overall, the \texttt{frALBERT}-based models have the lowest carbon footprint. These models offer a decrease of carbon emission between 20\% and 63\% compared to other models, depending on models and corpora. 
Note that \texttt{Carbon tracker} does not consider the execution environment or energy production. 
As a result, the obtained measures in our experiments remain approximative. \citet{touchent2023camembert} reported that the carbon emissions for pre-training their \texttt{CamemBERT-bio} model is estimated to 0.84 kg CO$_2$ eq. \citet{labrak-etal-2023-drbert} reported the overall carbon emissions of their 7 \texttt{DrBERT}-based models, which is 376.45 kg CO$_2$ eq.

\section{Conclusion}
This is the first benchmark evaluation of masked language models for the biomedical domain on the clinical French NER task, using three publicly available clinical French corpora. 
The evaluation is based on released gold standard annotations, including nested entities. 
\texttt{CamemBERT-bio} consistently outperforms \texttt{DrBERT}, while \texttt{FlauBERT} offers competitive results. Overall, \texttt{frALBERT} offers a fair compromise between \textit{F-measure} and carbon impact, with performance that exceeds the baseline consistently by at least 10 points, and a carbon impact that consistently represents a fraction of the impact of other models.  On the QUAERO French Med corpus, MLMs fail to outperform the knowledge-based approach proposed by \citet{erasmusCLEFeHealth2016}. This systematic evaluation compares model performance using metrics covering both performance and environmental impact. 

\section{Ethical considerations and limitations}
\paragraph{Limitations.} We did not consider all versions of models: for example, \texttt{camembert-large} and \texttt{flaubert-base-cased} are not covered in our experiments. While a full evaluation could cover more models, it would incur a higher carbon footprint and we decided to select representative models of the categories that we aimed to cover: general and domain-specific models that had been recently evaluated on similar corpora without direct comparability. 

  
\section{Acknowledgements}

We thank Dr. Bastien Rance for fruitful discussions on the content of this manuscript. This work was supported by ITMO-Cancer and ANR under grant CODEINE ANR-20-CE23-0026-01.  

\section{Bibliographical References}\label{sec:reference}

\bibliographystyle{lrec-coling2024-natbib}
\bibliography{biblio}

\section{Language Resource References}\label{lr:ref}
\bibliographystylelanguageresource{lrec-coling2024-natbib}
\bibliographylanguageresource{languageresource}

\end{document}